\documentclass{article} 
\usepackage{iclr2027_conference,times}

\usepackage{amsmath,amsfonts,bm}

\def\eqref#1{equation~\ref{#1}}

\def\1{\bm{1}}

\DeclareMathAlphabet{\mathsfit}{\encodingdefault}{\sfdefault}{m}{sl}
\SetMathAlphabet{\mathsfit}{bold}{\encodingdefault}{\sfdefault}{bx}{n}

\usepackage{graphicx}
\graphicspath{{figs/}}
\usepackage{booktabs}
\usepackage{multirow}
\usepackage{amssymb}
\usepackage{xcolor}
\usepackage{colortbl}          
\definecolor{cbest}{RGB}{178, 222, 171}    
\definecolor{csecond}{RGB}{222,238,222}  
\newcommand{\best}[1]{\cellcolor{cbest}\textbf{#1}}
\newcommand{\second}[1]{\cellcolor{csecond}\underline{#1}}
\newcommand{\bestbox}[1]{\colorbox{cbest}{\textbf{#1}}}        
\newcommand{\secondbox}[1]{\colorbox{csecond}{\underline{#1}}}
\usepackage{algorithm}
\usepackage{algpseudocode}
\definecolor{calg}{RGB}{53,120,200}
\algrenewcommand{\algorithmicrequire}{\textbf{Input:}}
\algrenewcommand{\algorithmiccomment}[1]{\hfill{\footnotesize\color{gray}$\triangleright$ #1}}
\newcommand{\algnew}{\textcolor{calg}{\textbf{+}}\ }
\usepackage{hyperref}
\usepackage{url}
\usepackage{pifont}

\newtheorem{proposition}{Proposition}

\title{Who Teaches Which Token? \\ Verifier-Gated Multi-Expert On-Policy Distillation for Scientific Reasoning}

\author{
Xun Xu\textsuperscript{1}\quad Zaixi Zhang\textsuperscript{2}\setcounter{footnote}{1}\thanks{Corresponding author. Email: \texttt{zaixizhang@ust.hk}.} \\
\textsuperscript{1}Fudan University \quad \textsuperscript{2}Hong Kong University of Science and Technology
}
\iclrfinalcopy 
\begin{document}

\maketitle
\lhead{}   
\renewcommand{\headrulewidth}{0pt}   
\suppressfloats[t]   

\begin{abstract}
Multi-teacher on-policy distillation (OPD) is becoming the standard way to
integrate specialist capabilities into one model: train experts with RL, then
distill them into the student on its own rollouts. Existing recipes assign
supervision at the sequence level --- each prompt goes to one domain teacher
and every token receives the same weight --- which implicitly assumes that a
teacher is uniformly useful across a response. We find instead that useful
teacher signal is sparse and heterogeneous along a reasoning trajectory, which
raises a finer question: \emph{who should teach which token?} \textbf{V}erifier-\textbf{G}ated Multi-Expert \textbf{O}n-\textbf{P}olicy \textbf{D}istillation (\textbf{VG-OPD}) answers it by verification: the
counterfactual gain of an expert on a specific answer criterion licenses that
expert to teach, its disagreement with the student localizes the supervision,
and criterion importance sets its weight; the gated KL
enters GRPO as an additive token-level advantage. Instantiated for scientific reasoning with RL-trained
capability experts, VG-OPD attains the best overall performance on seven
benchmarks for 4B and 8B students, ranking first on five at both scales,
with the largest gains on knowledge-intensive scientific reasoning tasks. Further analysis shows that the gains come from localizing verified
supervision rather than from adding teachers or distillation loss:
misplacing the same supervision budget is the single most damaging change,
and indiscriminate distillation drags RL below its own floor where gated
distillation lifts it.
\end{abstract}

\section{Introduction}
\label{sec:intro}

\begin{figure}[t]
\centering
\includegraphics[width=\linewidth]{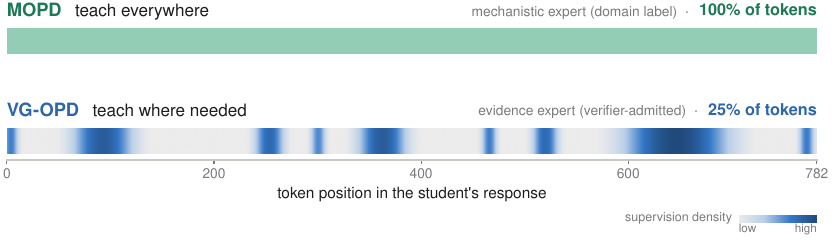}
\caption{\textbf{Who teaches which token?}
Token-level supervision comparison on one 782-token student rollout.
MOPD applies dense supervision from a single domain teacher, while VG-OPD
selectively assigns experts and concentrates supervision on tokens where the
selected expert disagrees with the student.
In this example, the evidence expert supervises only a subset of tokens,
with darker spans indicating stronger supervision.}
\label{fig:teaser}
\end{figure}

Scientific reasoning often requires combining multiple capabilities, such as
quantitative calculation, symbolic derivation, mechanistic explanation, and
evidence assessment, within a single answer. Reinforcement learning (RL) with verifiable or rubric rewards
provides a way to develop these capabilities using their respective training
data and verification procedures \citep{shao2024deepseekmathpushinglimitsmathematical,gunjal2025rubricsrewardsreinforcementlearning,chen2026improvingdatarewarddesign}.
To combine separately trained specialists into a single student,
multi-teacher on-policy distillation (OPD) supervises the student's own
rollouts with expert token probabilities
\citep{agarwal2024onpolicydistillationlanguagemodels,lu2025onpolicydistillation,ma2026mopdmultiteacheronpolicydistillation}.
However, how to route each expert's supervision to the parts of a student rollout where it is actually useful remains an open problem.

An expert's advantage can vary across both problems and requirements within a
problem. For example, on a chemistry question, an expert may correct the
student's calculation while giving an incorrect reaction mechanism.
MOPD \citep{ma2026mopdmultiteacheronpolicydistillation} assigns teachers by prompt-level domain labels and
applies uniformly weighted supervision across the response, which cannot
distinguish useful expert knowledge from erroneous reasoning within the same
answer. More recent methods improve distillation selectivity by gating
supervision with verifier rewards \citep{akhondzadeh2026rewardgatedonpolicydistillation,xu2026sgopdsigngatedonpolicydistillation}, selecting important tokens
\citep{xu2026tiptokenimportanceonpolicy,li2026croptaskrelevancecounterfactuals}, or routing supervision through additional signals
\citep{yu2026dopddualonpolicydistillation,wu2026dapddualanchoredpolicydistillation,xia2026cripoenhancingrubricbasedrl}. However, they determine where or from
which source to distill without explicitly verifying whether a candidate
expert has a criterion-specific advantage over the student. Selectivity alone does not establish eligibility: effective capability
integration requires deciding both which expert may teach and where that
expert's supervision should apply---\emph{who should teach which token?}

Our key idea is to decide teaching eligibility through a shared
criterion-level evaluation. Reference solutions decompose into \emph{criteria},
each paired with a verifier. Scoring expert and student
answers with the same criterion verifier identifies experts that are better
suited for specific requirements of the current problem. An expert should
therefore teach only when it shows a criterion-specific advantage over the
student, and only where it disagrees with the student's response:
verification establishes who may teach, and disagreement decides which tokens to teach (Fig.~\ref{fig:teaser}).

To implement this principle, we introduce \textbf{Verifier-Gated Multi-Expert
On-Policy Distillation (VG-OPD)}. \emph{Who}: VG-OPD evaluates
candidate experts under criterion-level verifiers and assigns supervision only
to experts that demonstrate an advantage over the student on the target
criterion. \emph{Which}: it identifies tokens where the selected expert
disagrees with the student and constructs a sparse supervision mask.

We instantiate VG-OPD with four RL-trained capability experts and evaluate
4B and 8B students on seven benchmarks spanning scientific reasoning, domain
science, and general reasoning. At 4B, VG-OPD
improves average accuracy over rubric-GRPO by $2.0$ points on scientific
reasoning and $3.3$ points on domain science, and consistently outperforms
coarse-grained multi-expert distillation (MOPD) across heterogeneous
scientific benchmarks. The 8B student shows the same trend, with VG-OPD
achieving the strongest overall performance across the evaluation suite. Ablations confirm that both verifier-based expert
eligibility and selective token supervision are critical to the gains.

\textbf{Contributions.}
\ding{182} We introduce a \emph{criterion-level eligibility principle} for multi-expert
distillation, where an expert is eligible to teach a criterion only when it
outperforms the student under the same verifier.
\ding{183} We propose \textbf{VG-OPD}, which operationalizes this principle through
verifier-gated expert selection and disagreement-based token supervision,
utilizing selective multi-expert distillation for RL training.
\ding{184} We provide controlled empirical evidence that \emph{both} criterion-level
expert eligibility and token-level supervision placement contribute to the
effectiveness of multi-expert distillation, with consistent improvements
across heterogeneous scientific reasoning benchmarks and model scales.

\section{Preliminaries}
\label{sec:setup}

\paragraph{Rubric-based RL yields failed criteria.}
For each prompt $x$, a rubric defines a set of verifiable criteria
$\mathcal{C}(x)=\{(c_j,V_j,\omega_j)\}_{j=1}^{J_x}$, where $c_j$ is a
criterion, $V_j(x,\cdot)\in[0,1]$ its verifier, and $\omega_j$ its signed
weight, negative for pitfall checks.
Following Rubrics-as-Rewards \citep{gunjal2025rubricsrewardsreinforcementlearning}, the rubric is reduced
to a scalar reward,
\begin{equation}
\label{eq:reward}
r(x,y)\;=\;\mathrm{clip}\!\Big(\frac{\sum_{j}\omega_j\,V_j(x,y)}{\sum_{j:\,\omega_j>0}\omega_j},\;0,\;1\Big),
\end{equation}
and GRPO \citep{shao2024deepseekmathpushinglimitsmathematical} optimizes it: a group of $G$ rollouts
$y^{(1)},\dots,y^{(G)}\sim\pi_{\theta_{\mathrm{old}}}(\cdot\mid x)$ is scored
with $r$, the group-normalized reward becomes a shared advantage for every
token of a rollout, and the policy takes a clipped step,
\begin{equation}
\label{eq:grpo}
\begin{aligned}
A^{\mathrm{task}}_{i}&=\frac{r(x,y^{(i)})-\mathrm{mean}_{g}\,r(x,y^{(g)})}{\mathrm{std}_{g}\,r(x,y^{(g)})},\\
\mathcal{L}_{\mathrm{PG}}(A;\theta)&=-\frac{1}{G}\sum_{i=1}^{G}\frac{1}{|y^{(i)}|}\sum_{t=1}^{|y^{(i)}|}
\min\!\Big(\rho_{i,t}A_{i,t},\ \mathrm{clip}\big(\rho_{i,t},1{-}\epsilon,1{+}\epsilon\big)A_{i,t}\Big),
\end{aligned}
\end{equation}
with $\rho_{i,t}$ the token importance ratio to $\pi_{\theta_{\mathrm{old}}}$ and
$A_{i,t}=A^{\mathrm{task}}_i$; GRPO minimizes
$\mathcal{L}_{\mathrm{PG}}(A^{\mathrm{task}};\theta)$. The scalar
discards structure the rubric provides for free: because $r$ is a sum over
criteria, every rollout comes with its set of \emph{failed criteria}
$\mathcal{F}(y)=\{j:\ \omega_j>0,\ V_j(x,y)=0\}$, and the same $V_j$ can
score any other answer to $x$ on the same criterion.
 
\paragraph{MOPD is fixed routing with dense weighting.}
On-policy distillation (OPD) matches the student to a teacher $E$ at every
generation step along the student's own rollouts
\citep{agarwal2024onpolicydistillationlanguagemodels,lu2025onpolicydistillation}. It yields a token-level
distillation signal $\hat k^{(E)}_t\ge0$, a single-sample estimate of the
per-position reverse KL, which enters the same
policy-gradient surrogate as the task reward as an advantage
$A^{\mathrm{KD}}_t=-w_t\hat k^{(E)}_t$, with $w_t\equiv1$ in plain OPD
\citep{ko2026scalingreasoningefficientlyrelaxed}. Multi-teacher OPD
merges $K$ RL-trained specialists $\{E_k\}$ into one student; MOPD
\citep{ma2026mopdmultiteacheronpolicydistillation} does so by assigning each prompt the expert of its domain
label $d(x)$ and weighting every token equally:
\begin{equation}
\label{eq:mopd}
\mathcal{L}_{\mathrm{MOPD}}(\theta)\;=\;\mathbb{E}_{x}\,\mathbb{E}_{y\sim\pi_\theta(\cdot\mid x)}
\Big[\sum_{t} w_t\,\hat k^{(E_{k_t})}_t\Big],\qquad k_t\equiv d(x),\quad w_t\equiv1.
\end{equation}
MOPD is therefore a special case of \emph{weighted token-level OPD}, indexed
by a teacher assignment $k_t$ and a weight field $w_t\in[0,1]$, in which
routing is fixed before the rollout is observed and all tokens receive equal
weight.

\paragraph{From fixed to verified assignment.}
The rubric says which criteria a rollout fails; the weighted OPD family says
who teaches which token and how much. VG-OPD keeps the objective of
Eq.~\ref{eq:mopd} and replaces its two fixed choices with verification-guided
teacher routing and sparse token weighting: for every failed criterion of
every rollout it decides \emph{which expert may teach}, \emph{which tokens}, and
\emph{with what strength}.

\section{VG-OPD: Verifier-Gated Multi-Expert On-Policy Distillation}
\label{sec:method}

\begin{figure}[t]
\centering
\includegraphics[width=\linewidth]{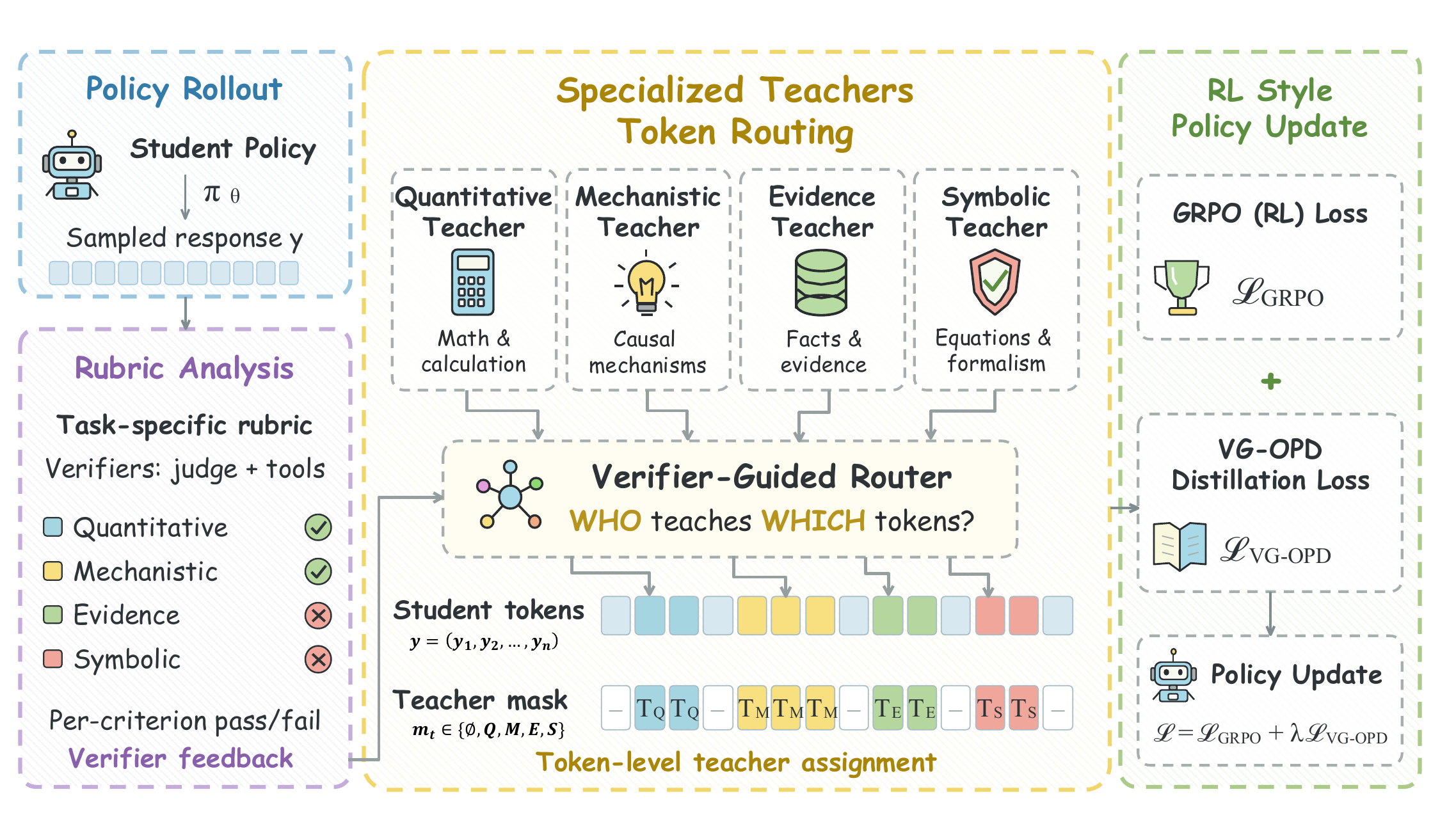}
\caption{Overview of \textbf{VG-OPD.} The student's rollouts are scored per criterion by the rubric's verifiers; for each failed criterion, the expert of that criterion's axis is licensed only if it answers better; the licensed teacher supervises only the tokens on which it disagrees with the student; the gated KL enters the GRPO update as an additive token-level advantage.}
\label{fig:pipeline}
\end{figure}

\subsection{Overview: who, where, and how much}
\label{sec:overview}
VG-OPD inserts one stage into the student's RL loop, between reward
computation and the policy update. For every rollout $y\sim\pi_{\theta_{\mathrm{old}}}(\cdot\mid x)$ and each
criterion it fails, we ask whether a candidate expert --- the expert of that
criterion's axis --- scores higher than the rollout under the criterion's
verifier. Only such experts teach (who), only on the tokens where they
disagree with the student (where), at a weight set by the criterion's rubric
importance (how much). The
decisions are per criterion, so a rollout that fails criteria on two axes
receives two teachers, each supervising its own token set, and the teacher
signal enters the update as a weighted token-level advantage added to the
task advantage.
An overview of the framework is in Fig.~\ref{fig:pipeline}. 

\subsection{Who teaches: licensing and routing}
\label{sec:gate}

\paragraph{Licensing.}
Each criterion and each expert carries a capability axis --- quantitative,
symbolic, mechanistic, or evidence --- written
$a(j)$ and $a(E_k)$. For a rollout $y$ with failed criteria $\mathcal{F}(y)$ and
$j\in\mathcal{F}(y)$, let $K(j)=\{k:a(E_k)=a(j)\}$
be the candidate experts of its axis. Each candidate produces a \emph{capability probe} $y_k=E_k(x)$: a plain greedy
answer to the original question, generated once per expert per prompt, cached,
and reused across all criteria and all $G$ rollouts of the group. The \emph{counterfactual gain} of expert $k$ on criterion $j$ --- the verifier
score its own answer receives on a criterion the rollout failed --- is
\begin{equation}
\label{eq:delta}
\Delta'_{j,k} \;=\; V_j\big(x,\,y_k\big) \;-\; V_j\big(x,\,y\big),
\end{equation}
and $k$ is \emph{licensed} to teach $c_j$ iff
$g_{j,k}:=\mathbf{1}[\Delta'_{j,k}>\delta]=1$. The gate measures, per criterion and per prompt, the condition
\citet{li2026rethinkingonpolicydistillationlarge} identify for OPD --- that the teacher offer a
capability the student lacks --- instead of assuming it: since
$V_j(x,y)=0$ on every failed criterion, the rollout enters the gate only
through $\mathcal{F}(y)$, and one cached verdict $V_j(x,y_k)$ serves all $G$
rollouts of the group. 
Three exclusions keep the gate honest:
pitfall criteria ($\omega_j<0$) never gate; verifier abstentions never gate;
and rollouts whose task reward is already high enough receive no teacher term.
A criterion with no licensed expert is simply not distilled this step ---
\emph{no teacher is better than an unverified teacher}. Weak experts are thereby rejected rather than averaged in.

\paragraph{Routing.}
\label{sec:route}
Among licensed candidates the router assigns
\begin{equation}
\label{eq:route}
q(k\mid x,y,c_j)\;=\;\frac{g_{j,k}\,\exp(\Delta'_{j,k}/\tau)}{\sum_{k'\in K(j)} g_{j,k'}\,\exp(\Delta'_{j,k'}/\tau)},\qquad
k^*(j)\;=\;\arg\max_{k\in K(j)}\ q(k\mid x,y,c_j),
\end{equation}
with temperature $\tau{=}0.5$; $q\equiv0$ when no candidate is licensed.
With one expert per axis, $K(j)$ is a singleton and $q$ reduces to the gate;
the softmax covers pools with several experts per axis. Licensing and
routing are per \emph{criterion}, not per rollout. Let
$\mathcal{L}(y)=\{j\in\mathcal{F}(y):\ g_{j,k^*(j)}=1\}$ be the licensed
criteria of a rollout; its teacher set is
$\mathcal{T}(y)=\{k^*(j):\ j\in\mathcal{L}(y)\}$. Because criteria of one
axis share that axis's expert, $|\mathcal{T}(y)|$ equals the number of
distinct axes among the licensed criteria. For example, a rollout with
licensed failures on the quantitative and the mechanistic axis is taught by
both experts, each on its own token set.

\subsection{Where to teach: disagreement localization}
\label{sec:localize}
Each teacher $E_k$, $k\in\mathcal{T}(y)$, restricts its distillation to a
token mask $m^{(k)}_t\in\{0,1\}$ over the student's own response, computed
under its own log-probabilities. Let
$\ell^{(k)}_t=\log\pi_{E_k}(y_t\mid x,y_{<t})$ be the teacher's
log-probability of the student's actual token and $Q_\eta(\ell^{(k)}_{1:|y|})$
the empirical $\eta$-quantile of these scores ($\eta{=}0.10$); then
\begin{equation}
\label{eq:mask}
\tilde m^{(k)}_t=\mathbf{1}\big[\ell^{(k)}_t\le Q_\eta(\ell^{(k)}_{1:|y|})\big],\qquad
m^{(k)}_t=\max\big(\tilde m^{(k)}_{t-1},\,\tilde m^{(k)}_t,\,\tilde m^{(k)}_{t+1}\big),
\end{equation}
i.e., the positions the teacher ``would not have written'', dilated by one
token on each side. The mask is rank-based, so it marks about $\eta$ of a response even
where the teacher largely agrees with it; that the teacher has something to
teach is guaranteed by the gate, not by the mask. Each mask comes from the
same forward pass of its teacher that supplies the distillation target, so
localization adds no calls.

\subsection{How much: verified-gain weights}
\label{sec:weight}
For each licensed $(\text{criterion},\text{teacher})$ pair, we assign

\begin{equation}
s_{j,k}=q(k\mid x,y,c_j)\,\bar\omega_j,
\end{equation}

where $q(k\mid x,y,c_j)$ denotes the verifier-derived teacher eligibility
and $\bar\omega_j=\big(\min(\omega_j,5)/5\big)^{1/2}\in(0,1]$ is the
criterion weight on the rubric's five-point scale after square-root
compression. The token-level supervision strength and teacher assignment are
then
\begin{equation}
\label{eq:weights}
w_t=\max_{j\in\mathcal{L}(y)}
s_{j,k^*(j)}\,m_t^{(k^*(j))},
\qquad
k_t=k^*(j_t),
\end{equation}
where $j_t$ is the criterion attaining the maximum with ties resolved in
favor of the teacher with larger total verified weight. Thus, each token is
assigned to at most one teacher, with its supervision determined by the
largest verified weight among the teachers licensed for the corresponding
criterion masks; overlapping teacher masks therefore do not produce
conflicting gradients. Each masked token is distilled from its selected
teacher alone, and tokens outside all masks receive no distillation gradient
($w_t=0$).

\paragraph{Objective.}
\label{sec:objective}
The teacher term is the weighted OPD of Eq.~\ref{eq:mopd}, using the
per-token teacher $k_t$ and sparse weight $w_t$ from Eq.~\ref{eq:weights}.
Its per-token signal is
\begin{equation}
\label{eq:opd}
\hat k_t=\min\!\left\{e^{\Delta_t}-\Delta_t-1,\;c\right\},\qquad
\Delta_t=\log\pi_{E_{k_t}}(y_t\mid x,y_{<t})
-\log\pi_\theta(y_t\mid x,y_{<t}),
\end{equation}
with $c=10$, whose expectation is the per-position reverse
Kullback--Leibler divergence $\mathrm{KL}(\pi_\theta\|\pi_{E_{k_t}})$.
The resulting advantage
$A^{\mathrm{KD}}_{i,t}=-w_{i,t}\hat k_{i,t}$ is detached from the gradient,
so the teacher signal simply suppresses tokens that the selected teacher
finds unlikely. The student is updated with two advantage streams:
\begin{equation}
\label{eq:joint}
\mathcal{L}(\theta)
=\mathcal{L}_{\mathrm{PG}}\!\left(A^{\mathrm{task}};\theta\right)
+\lambda\,\mathcal{L}_{\mathrm{PG}}\!\left(A^{\mathrm{KD}};\theta\right).
\end{equation}
Thus, GRPO reinforces task-level progress, while gated distillation
provides targeted teacher pressure where needed. Joint training is
important because $A^{\mathrm{KD}}\le0$ everywhere (Prop.~\ref{prop:sign},
App.~\ref{app:sign}); applying it alone can
continually suppress probability mass and accelerate entropy collapse.
The task-reward stream counterbalances this pressure, while the sparse
gate limits it to selected tokens. Alg.~\ref{alg:vgopd} lists one training
step (implementation in App.~\ref{app:impl}); lines marked {\color{calg}\textbf{+}} are the
additions to a standard GRPO step.

\begin{algorithm}[t]
\caption{VG-OPD training step}
\label{alg:vgopd}
\small
\begin{algorithmic}[1]
\Require prompt $x$ with criteria $\mathcal{C}(x)$, student $\pi_\theta$, experts $\{E_k\}$
\State sample $\{y^{(i)}\}_{i=1}^G\sim\pi_{\theta_{\mathrm{old}}}(\cdot\mid x)$; score every criterion $V_j(x,y^{(i)})$ \Comment{GRPO}
\State $A^{\mathrm{task}}\gets$ group-normalized task rewards \Comment{GRPO}
\State \algnew $y_k\gets E_k(x)$ for each expert $k$ (greedy probe, cached per prompt) \Comment{who: probe}
\State \algnew \textbf{for} each rollout $y$ and each failed criterion $j\in\mathcal{F}(y)$ \textbf{do}
\State \algnew \hspace{\algorithmicindent} $\Delta'_{j,k}\gets V_j(x,y_k)-V_j(x,y)$ for $k\in K(j)$ \Comment{Eq.~\ref{eq:delta}}
\State \algnew \hspace{\algorithmicindent} license $k$ iff $\Delta'_{j,k}>\delta$; route $k^*(j)$ by Eq.~\ref{eq:route}; set $s_{j,k^*(j)}$ \Comment{gate, route}
\State \algnew \textbf{for} each licensed teacher $k\in\mathcal{T}(y)$ \textbf{do}
\State \algnew \hspace{\algorithmicindent} one forward pass of $E_k$ over $y$ gives $\ell^{(k)}_t$; mask $m^{(k)}_t$ by Eq.~\ref{eq:mask} \Comment{where}
\State \algnew $w_t,\,k_t\gets$ Eq.~\ref{eq:weights} \Comment{how much: one teacher per token}
\State \algnew $A^{\mathrm{KD}}_t\gets -w_t\,\hat k_t$ under $E_{k_t}$, no gradient through $\hat k_t$ \Comment{Eq.~\ref{eq:opd}}
\State update $\theta$ on $\mathcal{L}_{\mathrm{PG}}(A^{\mathrm{task}})+\lambda\,\mathcal{L}_{\mathrm{PG}}(A^{\mathrm{KD}})$ \Comment{Eq.~\ref{eq:joint}; GRPO}
\end{algorithmic}
\end{algorithm}

\section{Experiments}
\label{sec:exp}

\subsection{Experimental Setup}
\label{sec:setup-exp}

\paragraph{Data and models.} Students and experts are built on Qwen3-4B and
Qwen3-8B \citep{yang2025qwen3technicalreport} in non-thinking mode. The four capability experts, one
per axis, are LoRA adapters on the shared base, each trained with standard
GRPO on its axis's slice of the training data (App.~\ref{app:experts});
training data, criteria bank, and verifiers are described in
Apps.~\ref{app:data}--\ref{app:verifiers}; training details are in
App.~\ref{app:training}. 

\paragraph{Baselines.} The starting checkpoint
(Vanilla); GRPO with rubric rewards \citep{gunjal2025rubricsrewardsreinforcementlearning}; OPSD, self-distillation
with the rubric as privileged context; MOPD \citep{ma2026mopdmultiteacheronpolicydistillation}; teacher-anchored GRPO (T.A.~GRPO)
\citep{ramos2026recipelongcontextreasoninglarge}; and CriPO \citep{xia2026cripoenhancingrubricbasedrl}. All distillation arms
share data, budget, and $\lambda$ and differ only in the teacher
(App.~\ref{app:baselines}). 

\paragraph{Evaluation benchmarks.} We evaluate on seven public
benchmarks in three groups: \textbf{scientific reasoning} (GPQA-Diamond, RaR-Science,
SciBench), \textbf{domain science} (ChemBench, RaR-Med), and \textbf{general reasoning}
(MMLU-Pro, MATH500). Benchmark details are in App.~\ref{app:benchmarks}. 

\begin{table}[t]
\centering
\caption{\textbf{Main results} (accuracy, \%) on seven benchmarks in three groups for the 4B and 8B students; ``distill.\ only'' is VG-OPD without the task-reward stream; T.A.~GRPO is the teacher-anchored GRPO proposed by \citet{ramos2026recipelongcontextreasoninglarge}. Within each block, the \bestbox{best} result per column is bold on dark green and the \secondbox{second-best} is underlined on light green.}
\label{tab:main}
\footnotesize
\setlength{\tabcolsep}{3pt}
\newsavebox{\maintabbox}
\begin{lrbox}{\maintabbox}
\begin{tabular}{l l ccc cc cc}
\toprule
& & \multicolumn{3}{c}{Scientific reasoning} & \multicolumn{2}{c}{Domain science} & \multicolumn{2}{c}{General reasoning} \\
\cmidrule(lr){3-5}\cmidrule(lr){6-7}\cmidrule(lr){8-9}
& Method & GPQA-D & RaR-Sci & SciBench & ChemBench & RaR-Med & MMLU-Pro & MATH500 \\
\midrule
\multirow{8}{*}{\rotatebox{90}{Qwen3-4B}}
 & Vanilla                  & 41.4 & 57.9 & 50.8 & 57.8 & 46.9 & 57.2 & 80.9 \\
 & GRPO                     & 43.9 & 60.6 & 54.5 & 61.2 & 50.7 & 62.5 & \best{85.1} \\
 & OPSD                     & 42.1 & 59.6 & 52.1 & 58.9 & 48.5 & 60.2 & 82.8 \\
 & MOPD                     & 43.8 & 59.0 & 54.0 & 63.1 & 49.4 & 61.6 & 82.4 \\
 & T.A. GRPO   & 45.5 & \second{61.2} & 51.9 & 62.0 & \second{51.0} & \best{63.8} & 83.6 \\
 & CriPO                    & \second{45.7} & 59.1 & 54.5 & \second{65.5} & 50.9 & 62.5 & 84.2 \\
 & VG-OPD (distill.\ only)  & 43.4 & 60.7 & \second{55.2} & 64.3 & 50.2 & 61.3 & 83.0 \\
 & \textbf{VG-OPD}          & \best{46.3} & \best{62.4} & \best{56.2} & \best{66.4} & \best{52.2} & \second{63.3} & \second{84.6} \\
\midrule
\multirow{8}{*}{\rotatebox{90}{Qwen3-8B}}
 & Vanilla                  & 41.9 & 59.7 & 55.2 & 64.6 & 53.8 & 64.1 & 83.6 \\
 & GRPO                     & 45.8 & 64.5 & 57.4 & \second{69.3} & 57.4 & \best{67.1} & \second{85.4} \\
 & OPSD                     & 42.2 & 60.6 & 55.9 & 66.5 & 55.4 & 64.9 & 84.2 \\
 & MOPD                     & 44.1 & \second{66.2} & 56.1 & 67.2 & 56.9 & 65.6 & 84.8 \\
 & T.A. GRPO    & 45.4 & 65.5 & 56.8 & 68.1 & 57.1 & 66.0 & 83.4 \\
 & CriPO                    & 45.8 & 64.8 & \second{57.6} & 66.7 & \second{57.8} & 65.2 & 84.2 \\
 & VG-OPD (distill.\ only)  & \second{46.3} & 65.9 & 57.2 & 67.0 & 56.2 & 65.8 & 84.5 \\
 & \textbf{VG-OPD}          & \best{47.3} & \best{67.0} & \best{59.1} & \best{70.9} & \best{60.1} & \second{66.7} & \best{85.7} \\
\bottomrule
\end{tabular}
\end{lrbox}
\ifdim\wd\maintabbox>\linewidth
  \resizebox{\linewidth}{!}{\usebox{\maintabbox}}
\else
  \usebox{\maintabbox}
\fi
\end{table}

\subsection{Main Results}
\label{sec:main}
\ding{182} \textbf{VG-OPD is best overall at both scales} (Table~\ref{tab:main}):
first on five of seven benchmarks and second, within half a point, on
MMLU-Pro and MATH500.
\ding{183} \textbf{The gains concentrate on knowledge-intensive science while
general reasoning is preserved.} Over rubric-GRPO, the 4B student adds $+2.0$
on scientific reasoning, $+3.3$ on domain science, and $+0.2$ on general
reasoning (group averages); the 8B student adds $+1.7$ and $+1.9$ and stays
within $0.3$ on general reasoning.
\ding{184} \textbf{Every other distillation scheme trails VG-OPD.} Pure distillation (OPSD, MOPD) at best ties rubric-GRPO on
group averages, and dense fusion (teacher-anchored GRPO) stays within half a point of the RL
floor at 4B and at or below it at 8B, by up to $1.6$ points.

\begin{table}[t]
\centering
\caption{\textbf{Training-paradigm comparison} (group averages, \%). The teacher pool is fixed; only the assignment granularity changes: a privileged self-teacher (OPSD), domain-routed experts (MOPD), or criterion-routed and gated experts (VG-OPD), used standalone or as auxiliary supervision inside GRPO. \bestbox{Best} per column is bold on green.}
\label{tab:paradigm}
\small
\begin{tabular}{l l ccc}
\toprule
Regime & Method & Sci.\ reasoning $\uparrow$ & Domain science $\uparrow$ & General reasoning $\uparrow$ \\
\midrule
\multirow{3}{*}{Standalone} & OPSD & 51.3 & 53.7 & 71.5 \\
 & MOPD                              & 52.3 & 56.3 & 72.0 \\
 & VG-OPD                            & 53.1 & 57.3 & 72.2 \\
\midrule
\multirow{3}{*}{RL + distill.} & GRPO + OPSD & 51.6 & 54.5 & 73.7 \\
 & GRPO + MOPD                       & 50.1 & 56.4 & 73.2 \\
 & \textbf{GRPO + VG-OPD}            & \best{55.0} & \best{59.3} & \best{74.0} \\
\midrule
RL only & GRPO                       & 53.0 & 56.0 & 73.8 \\
\bottomrule
\end{tabular}
\end{table}

\begin{table}[t]
\centering
\caption{\textbf{Component-wise ablation} (group averages, \%). Each variant disables exactly one component of the token-weight construction (Eq.~\ref{eq:weights}); data, steps and $\lambda$ are unchanged. \bestbox{Best} per column is bold on green.}
\label{tab:component}
\small
\begin{tabular}{l l ccc}
\toprule
Component & Variant & Sci.\ reasoning $\uparrow$ & Domain science $\uparrow$ & General reasoning $\uparrow$ \\
\midrule
 & \textbf{Full VG-OPD} & \best{55.0} & \best{59.3} & \best{74.0} \\
\midrule
\multirow{2}{*}{Teacher assignment} & w/o gate      & 50.6 & 57.7 & 73.5 \\
 & Random routing                                    & 49.3 & 55.3 & 70.2 \\
\midrule
\multirow{3}{*}{Localization} & Sequence-level      & 52.5 & 56.0 & 73.3 \\
 & Step-level                                        & 51.5 & 56.5 & 71.2 \\
 & Random localization                               & 48.7 & 47.2 & 68.4 \\
\midrule
Weighting & Uniform weighting                        & 51.4 & 57.2 & 73.2 \\
\bottomrule
\end{tabular}
\end{table}

\subsection{Ablation Studies}
\label{sec:ablation}
\textbf{Training paradigm (Table~\ref{tab:paradigm}).} With the teacher pool
fixed, finer assignment wins in both regimes. Standalone, accuracy rises from
OPSD to MOPD to VG-OPD (scientific reasoning $51.3\to52.3\to53.1$, domain
science $53.7\to56.3\to57.3$), and standalone VG-OPD matches or exceeds the
RL-only floor on both science groups without any task reward. Inside GRPO the contrast sharpens:
dense distillation drags GRPO \emph{below its own floor} (GRPO+OPSD $51.6$ and
GRPO+MOPD $50.1$ vs.\ $53.0$ on scientific reasoning), whereas gated,
localized distillation lifts it by $+2.0$ / $+3.3$ / $+0.2$. Verifying and
localizing the teacher signal is thus what turns distillation into a gain on
top of RL.

\textbf{Components (Table~\ref{tab:component}).} Each variant disables one
component of Eq.~\ref{eq:weights}, with data, steps, and $\lambda$ unchanged.
Localization matters most: a coverage-matched random mask loses $6.3$ /
$12.1$ / $5.6$ points (scientific / domain / general), more than a
sequence-level ($2.5$ / $3.3$ / $0.7$) or step-level mask ($3.5$ / $2.8$ /
$2.8$), so distilling the same budget at the wrong positions is worse than
distilling everywhere. Who teaches matters as well: random routing costs
$5.7$ / $4.0$ / $3.8$ and removing the gate $4.4$ / $1.6$ / $0.5$, both
largest on scientific reasoning, and flat weights cost $3.6$ / $2.1$ / $0.8$.
App.~\ref{app:comparisons} plots both tables as differences.

\begin{figure}[t]
\centering
\begin{minipage}[t]{0.44\linewidth}\centering\includegraphics[width=\linewidth]{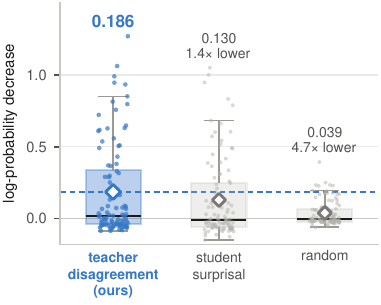}\\{\small (a) where: log-probability decrease at three token sets}\end{minipage}\hfill
\begin{minipage}[t]{0.52\linewidth}\centering\includegraphics[width=\linewidth]{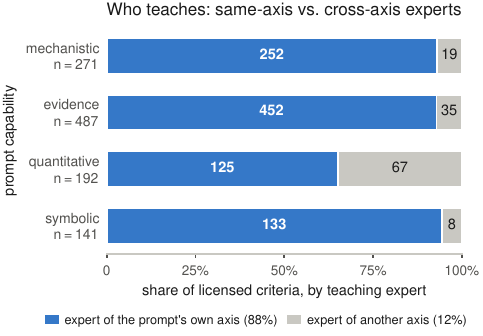}\\{\small (b) who: teaching expert per prompt capability}\end{minipage}
\caption{\textbf{Mechanism analysis.} (a) Decrease $D_t$ in the starting model's log-probability of its own tokens after training, per rollout, at teacher-disagreement, highest-surprisal, and random positions (box: IQR, line: median, diamond: mean). (b) Who teaches the licensed criteria: the prompt's own-axis expert (blue) or another axis's expert (grey) with counts in the bars.}
\label{fig:signal}
\end{figure}
\subsection{Framework Analysis}
\label{sec:signal}

\paragraph{Where: the mask selects corrective tokens.} Let
$D_t=\log p_{\mathrm{start}}(y_t\mid x,y_{<t})-\log p_{\mathrm{train}}(y_t\mid x,y_{<t})$
be the decrease, after training, in the starting model's log-probability of its
own token. Averaged per rollout over equal-coverage token sets
(Fig.~\ref{fig:signal}a), $D_t$ is $0.186$ at teacher-disagreement positions,
$0.130$ at the student's highest-surprisal positions, and $0.039$ at random
positions. The effect is concentrated: the teacher-marked tokens drop by at
least $0.05$ in $43\%$ of rollouts, and in $98\%$ of these they drop more
than random positions; over all rollouts they drop more than the
highest-surprisal positions in $85\%$. The mask thus concentrates updates on
teacher-identified disagreements rather than merely on uncertain tokens
\citep{xu2026tiptokenimportanceonpolicy,wang2026disagreementlearnabletokenteachability}. The localization ablations in
Table~\ref{tab:component} complete the picture: a sequence-level, step-level,
or coverage-matched random mask lowers accuracy, and the random mask, which
keeps the coverage, lowers it most, so the gain depends on \emph{where} the
signal is applied, not on how many tokens receive it.

\paragraph{Who: routing is specialization-aligned, not hard-coded.}
$88\%$ of licensed criteria are taught by the expert of the prompt's own
capability (Fig.~\ref{fig:signal}b), so routing recovers the specialization
that MOPD imposes by label. The remaining $12\%$ cross axes (symbolic
criteria of quantitative prompts, mechanistic criteria of evidence prompts)
show that the mapping from prompt to teacher is not fixed: a prompt's
failed criteria can call in a second expert.

\begin{figure}[t]
\centering
\includegraphics[width=\linewidth]{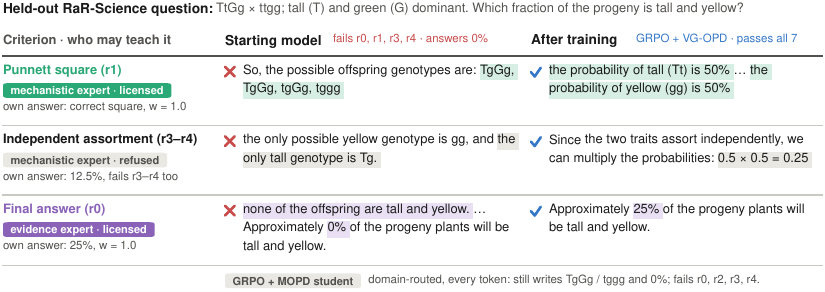}
\caption{\textbf{Who teaches which failure.} Answers of the starting model and of the GRPO$+$VG-OPD trained model on a held-out RaR-Science question (verbatim excerpts).}
\label{fig:textcase}
\end{figure}

\subsection{Case Study}
\label{sec:case}
Fig.~\ref{fig:textcase} traces the procedure on a held-out RaR-Science
question whose starting answer fails four criteria. Each failed criterion is
put to the expert of its axis, and that expert's own answer decides: the
mechanistic expert is licensed on the criterion it satisfies and refused on
the two it fails itself, which therefore receive no teacher this step, while
the evidence expert is licensed on the final-answer criterion. Each licensed
expert then supervises only the tokens on which it disagrees with the
student. After training, the GRPO$+$VG-OPD student passes every criterion,
whereas the GRPO$+$MOPD student, distilled from the domain expert on every
token, still fails the answer. App.~\ref{app:cases} gives two further cases.

\section{Related Work}
\label{sec:related}

\paragraph{Multi-teacher on-policy distillation.}
On-policy distillation (OPD) trains a student on its own rollouts under
per-token teacher supervision \citep{agarwal2024onpolicydistillationlanguagemodels,lu2025onpolicydistillation},
and its multi-teacher form is now the standard recipe for merging specialists
trained by RL. MOPD \citep{ma2026mopdmultiteacheronpolicydistillation} routes each prompt to the teacher of
its domain label and applies uniform sampled-token reverse KL on every token;
H-OPD arbitrates among teachers token by token by teacher confidence \citep{yin2026hopdconfidenceawareheterogeneous}, TU-OPD
gates whole examples by reliability scores \citep{lu2026bestteacherexpandingcompressing}, and CaMOPD
addresses recovery--preservation conflicts between teachers
\citep{chen2026counteractionawaremultiteacheronpolicydistillation}. \citet{li2026rethinkingonpolicydistillationlarge} show that a teacher helps only
where it offers a capability the student lacks.

\paragraph{Selective and localized distillation.}
Trust gates decide \emph{when} a teacher may speak: RG-OPD keeps a rollout
only when its verifier reward and the teacher--student likelihood gap agree
\citep{akhondzadeh2026rewardgatedonpolicydistillation}, RSTG distills only on
prompts whose rollouts all fail \citep{han2026distillfailrecoveringlearning},
SG-OPD gates each token by whether the teacher agrees with the
verifier-correct direction \citep{xu2026sgopdsigngatedonpolicydistillation},
SGSD validates the polarity of each skill-conditioned self-teacher against
the rollout outcome \citep{huang2026skillconditionedgatedselfdistillationllm},
and TrOPD confines the update to regions where the teacher is reliable
\citep{xing2026trustregiononpolicydistillation}. Token selectors decide
\emph{where}: decision-critical tokens missed by top-$K$ truncation
\citep{shen2026topkmissesdecisiontoolcall}, token importance and teachability
\citep{xu2026tiptokenimportanceonpolicy,wang2026disagreementlearnabletokenteachability},
counterfactual task relevance \citep{li2026croptaskrelevancecounterfactuals},
supervision routed between privileged teacher and student by advantage gap
and relative probability \citep{yu2026dopddualonpolicydistillation} or
anchored to reference paths \citep{wu2026dapddualanchoredpolicydistillation},
and outcome-contrasted patches in Woodpecker
\citep{wang2026woodpeckerdistillationweakmodels}. All of these gate or select
against a single task-level verdict or an optimization statistic; none
checks, per criterion, whether the teacher itself satisfies the criterion it
is asked to teach.

\paragraph{Rubric-based RL for science.}
Rubric rewards underpin recent scientific RL \citep{gunjal2025rubricsrewardsreinforcementlearning,chen2026improvingdatarewarddesign}, with
criterion-to-verifier routing for robust rewards \citep{yu2026reinforcementlearningrobustrubric} and
automated rubric synthesis \citep{li2026aresautomatedrubricsynthesis,guan2026evorubricselfevolvingrubricdrivenrl}. Rubrics also serve
as privileged teacher context in self-distillation
\citep{gu2026rethinkingrewardsupervisionrubricconditioned,bablani2026rubricsprivilegedinformationopenended,rezaei2026rubricguidedselfdistillationposttrainingrubric}
or replace teacher logits as the scoring signal of on-policy distillation
\citep{fang2026rubricbasedonpolicydistillation}, and CriPO decomposes
rubric-RL by criterion and self-distills criterion-conditioned revisions with
token localization \citep{xia2026cripoenhancingrubricbasedrl}. 
\section{Conclusion}
\label{sec:conclusion}
Multi-teacher on-policy distillation asks which teacher to use; we argued that
it should ask who teaches which token. VG-OPD answers by verification: the counterfactual gain of an expert on a
specific criterion licenses, routes, and weights the teacher signal, and the
expert's disagreement with the student localizes it, all inside a standard
GRPO update. On scientific reasoning it outperforms uniform multi-teacher distillation,
privileged self-distillation, dense fusion, and rubric-based RL, trains to
completion where dense distillation does not, and owes its gains to
\emph{where} it teaches rather than how much. We expect the same question
to matter wherever specialist teachers are merged.

\section*{Limitations}
This work studies VG-OPD on a broad but still finite set of scientific
benchmarks and student sizes. The seven benchmarks are text-only questions
concentrated on physics, chemistry, biology, medicine, and mathematics; other
scientific domains, and modalities such as figures, molecular structures,
and experimental data, are not covered and would require extending the
criteria bank and its verifiers. The students are 4B and 8B models; although
the results suggest that stronger students may benefit from the same
verified, localized teaching, this scaling behavior remains to be verified
systematically.

\subsection*{AI use statement}
In this work, we used generative AI tools to improve the readability of the manuscript and to provide auxiliary coding support. All AI-assisted manuscript edits were reviewed by the authors, and AI-assisted code was verified and tested by the authors. We take responsibility for the final content of this work, including all text, claims, and artifacts produced with the aid of generative AI.

\subsection*{Reproducibility statement}
The implementation of VG-OPD, including the training and evaluation code and configuration files, will be released publicly. The training data, criteria bank, and verifiers (Apps.~\ref{app:data}--\ref{app:verifiers}), expert training (App.~\ref{app:experts}), the implementation (App.~\ref{app:impl}), training details (App.~\ref{app:training}), benchmark details (App.~\ref{app:benchmarks}), and baseline construction (App.~\ref{app:baselines}) are documented in the appendix. 

\bibliography{refs}
\bibliographystyle{iclr2027_conference}

\appendix
\section{Method Details}
\label{app:method}

\subsection{Sign and Scale of the Teacher Stream}
\label{app:sign}
The distillation advantage $A^{\mathrm{KD}}_t=-w_t\hat k_t$ of Eq.~\ref{eq:opd}
is non-positive everywhere, so on its own it only suppresses probability mass;
this is why Eq.~\ref{eq:joint} keeps the task-reward stream and why the gate
and the mask matter for stability.
\begin{proposition}[sign and scale of the teacher stream]
\label{prop:sign}
For every position $t$, $e^{\Delta_t}-\Delta_t-1\ge0$, with equality iff
$\pi_E(y_t\mid x,y_{<t})=\pi_\theta(y_t\mid x,y_{<t})$, and
$\mathbb{E}_{y_t\sim\pi_\theta(\cdot\mid x,y_{<t})}[e^{\Delta_t}-\Delta_t-1]=\mathrm{KL}_t$.
Hence $A^{\mathrm{KD}}_t\le0$ pointwise, $\mathbb{E}[A^{\mathrm{KD}}_t]=-w_t\,\mathrm{KL}_t$
for a fixed weight before clamping, and the expected teacher pressure on a
rollout obeys $\sum_t w_t\,\mathrm{KL}_t\le\bar s\sum_{t:\,w_t>0}\mathrm{KL}_t$ with
$\bar s=\max_j s_{j,k^*(j)}\le1$, against $\sum_t\mathrm{KL}_t$ under MOPD.
\end{proposition}

\paragraph{Proof.}
Non-negativity is $e^{u}\ge1+u$ with equality iff $u=0$, applied to
$u=\Delta_t$. For the expectation,
$\mathbb{E}_{y_t\sim\pi_\theta}[e^{\Delta_t}]=\sum_v\pi_\theta(v)\,\pi_E(v)/\pi_\theta(v)=1$
because softmax policies have full support, and
$\mathbb{E}_{y_t\sim\pi_\theta}[\Delta_t]=-\mathrm{KL}_t$ by definition of the
reverse KL, so $\mathbb{E}[e^{\Delta_t}-\Delta_t-1]=1+\mathrm{KL}_t-1=\mathrm{KL}_t$.
The sign of $A^{\mathrm{KD}}_t=-w_t\hat k_t$ follows from $w_t\ge0$, and the
pressure bound from $w_t\le\bar s$ on the mask and $w_t=0$ elsewhere; the clamp
$c$ only lowers $\hat k_t$ and leaves the sign unchanged. \hfill$\square$

\subsection{Implementation Details}
\label{app:impl}

\paragraph{Trainer integration.}
We build on the distillation stack of \emph{verl} \citep{Sheng_2025} (teacher
log-probabilities streamed per token; sampled-token reverse-KL with an
optional policy-gradient form; multi-teacher routing by a per-sample key).
VG-OPD adds: (i) a registered distillation loss that multiplies the per-token
KL estimate by the per-token weights $w_t$ --- absent weights fall back to all
ones, so every baseline shares one loss implementation; (ii) one hook in the
rollout worker, placed after reward computation and before teacher scoring,
which parses per-criterion results, runs probes/gate/routing/localization
asynchronously per sample, and attaches the weight vector and, for every
token, the log-probability under that token's teacher to the batch; the hook is fail-open (any internal
error zero-fills weights rather than stalling training). Rewards, probe
verdicts, and judge calls are asynchronous and overlap rollout generation;
measured reward wall-time per step is effectively zero.
\section{Experimental Details}
\label{app:setup}

\subsection{Training Data}
\label{app:data}
Training prompts, reference answers, and open-ended rubrics derive from the Dr.SCI collection
\citep{chen2026improvingdatarewarddesign} (via a publicly available reproduction of its pipeline, since the official release was
unavailable at the time of writing). Although we use it as one mixture, it aggregates four datasets with
different task forms. \emph{WebInstruct-Verified} \citep{ma2025generalreasoneradvancingllmreasoning}: web-sourced questions
across physics, chemistry, mathematics, and other disciplines, filtered to those whose short answer (a
number, an expression, or a multiple-choice letter) can be checked automatically. \emph{NaturalReasoning}
\citep{yuan2025naturalreasoningreasoningwild28m}: open-ended reasoning questions extracted from pretraining corpora and paired
with model-written reference answers, spanning many domains. \emph{MegaScience} \citep{fan2025megasciencepushingfrontiersposttraining}: a
large mixture of textbook-derived and existing scientific question--answer data across physics, chemistry,
biology, medicine, computer science, economics, and mathematics, with reference answers in both short-answer and
free-form styles. \emph{RaR-Science} \citep{gunjal2025rubricsrewardsreinforcementlearning}: open-ended science questions paired with expert-guided
rubrics. Closed-form items enter the derive-then-verify route of App.~\ref{app:criteria} and open-ended items
the rubric-import route. Licenses: WebInstruct-Verified (Apache-2.0), NaturalReasoning (CC-BY-NC-4.0),
MegaScience (CC-BY-NC-SA-4.0), and RaR-Science (no explicit dataset license; upstream includes non-commercial
sources); the mixture is therefore non-commercially encumbered, and we use it for research only, do not
relicense it, and release derived criteria banks under the most restrictive upstream terms. Evaluation sets
are public benchmarks used under their own licenses.

\subsection{Criteria Bank}
\label{app:criteria}
\paragraph{Sources and derive-then-verify construction.}
Criteria come from two routes. (i) \emph{Verifiable} items (numeric,
expression, MCQ): reference answers are typically bare final values (median
${\sim}11$ characters), so intermediate criteria cannot be extracted from
them. Instead the annotation model solves the problem itself; its derivation
is admitted \emph{only if} its final answer passes an endpoint check against
the ground truth, after which checkpoint steps become intermediate criteria
(restricted to deterministically verifiable types, deduplicated, with
target-parsability checks). (ii) \emph{Open-ended} items: the source rubrics
are imported verbatim --- zero rewriting, audit-friendly --- then gated for
usability, per-item validity, and axis assignment; style/organization items
(no capability axis) are dropped; per axis we keep the top-6 items by
$|\omega|$ with the primary axis first. Weights keep their sign; ${\sim}16\%$
of stored criteria are penalty (pitfall) items. A row is dropped unless its
primary axis has at least one positive-weight criterion (no positive learning
signal otherwise).

\paragraph{Task reward aggregation.}
The scalar task reward of Eq.~\ref{eq:reward} is computed per capability axis as
the signed, weight-normalized sum of criterion scores clipped to $[0,1]$,
$r_a=\mathrm{clip}\big(\sum_j \omega_j V_j/\sum_{j:\omega_j>0}\omega_j,0,1\big)$:
positive-weight criteria contribute their graded verifier score, pitfall
criteria subtract $|\omega_j|$ when triggered, and an axis with no
positive-weight criterion scores $0$. Verifier abstentions (\texttt{unknown},
including criteria left unscored when a sample exhausts its verifier time
budget) are excluded from both numerator and normalizer. The final reward
multiplies the axis score by a binary format term (a boxed final answer is
required), and expert training uses the reward of the expert's own axis.

\subsection{Verifiers}
\label{app:verifiers}
\paragraph{Verifier registry.}
\texttt{numeric+unit} (parser + unit conversion + relative tolerance),
\texttt{sympy\_equiv} (LaTeX$\to$symbolic equivalence; timeout returns
\texttt{unknown}, which never enters rewards or the gate),
\texttt{sandbox\_py} (resource-limited local subprocess re-computation),
\texttt{rule} (constraint/normalized-match checks), and \texttt{llm\_judge}
for mechanistic/evidence criteria. Judge calls are cached on
(judge version, prompt version, criterion id, answer hash); the (model,
version, prompt version) triple is recorded per verdict. Rollouts must end in
a boxed final answer; format failures zero the format term but open-ended
content scoring is not gated on it.

\paragraph{Judge calibration.}
We calibrate the judge by cross-model referee: a
stratified sample of (criterion, answer) pairs --- two judge axes $\times$
four subjects $\times$ positive/penalty items, with answers drawn from own
reference (should pass), same-subject mismatched reference (fluent hard
negative), and vague template (should fail) --- is scored by the production
judge under the production protocol, independently by a stronger cross-family
referee with full context, and disagreements are adjudicated blind by a third
model. Outcomes ($n{=}578$): overall agreement $87.0\%$; positive-weight items
$92.8\%$; per-axis precision $0.877$ (mechanistic) / $0.782$ (evidence); vague answers are never
credited ($0/100$), and fluent mismatched references are rejected at $92\%$.
The salient weakness is penalty items: $39\%$ false-trigger rate on clean
answers (the judge over-attributes errors of omission), which is why pitfall
criteria are excluded from gating (\S\ref{sec:gate}) and carry small
$|\omega|$ in rewards.

\subsection{Capability Experts}
\label{app:experts}

Each expert is a LoRA adapter (rank $64$) on the shared base, trained with standard
GRPO on the axis-filtered slice of the training data against its own axis
reward: the quantitative and symbolic experts on deterministic verifier
rewards, the mechanistic and evidence experts on graded judge rewards over the
imported rubrics. Runs are capped in steps and stopped on sustained validation
decline, and the deployed checkpoint of each expert is selected by offline
re-scoring on the internal dev set.

\subsection{Training Details}
\label{app:training}
Expert training is described in App.~\ref{app:experts}; this section gives
the student's configuration and checkpoint selection. Table~\ref{tab:hparams} lists the configuration of the VG-OPD runs, used
unchanged for the 4B and the 8B student. All baselines share it and differ
only in teacher construction (App.~\ref{app:baselines}); the RL-only floor differs only in the entropy bonus
(set to $0$, for the reason given there); the ablation arms of
Table~\ref{tab:component} change exactly one switch each.
\begin{table}[h]
\centering
\small
\setlength{\tabcolsep}{5pt}
\caption{Training and evaluation configuration of the VG-OPD runs (4B and 8B students).}
\label{tab:hparams}
\begin{tabular}{l p{3.9in}}
\toprule
Student & Qwen3-4B, Qwen3-8B; LoRA rank $64$, $\alpha{=}32$, all linear layers \\
Data per step & $21$ prompts $\times$ $8$ rollouts ($168$ responses); prompt $\le1{,}024$ tokens; response $\le4{,}096$ tokens; context $5{,}120$ \\
Optimizer & AdamW, learning rate $10^{-5}$; one PPO epoch and one minibatch per step; PPO clip $0.2$; dynamic batching at $6{,}144$ tokens per GPU \\
GRPO & group-normalized advantages; entropy bonus $0.01$; reference-KL penalty $10^{-3}$ (low-variance estimator) \\
Distillation & $\lambda{=}0.25$ (Eq.~\ref{eq:joint}); sampled-token teacher log-probability with the estimator $\hat k_t$ of Eq.~\ref{eq:opd} ($e^{\Delta}-\Delta-1$, clamped at $c{=}10$; verl's $k_3$ estimator), used as a detached advantage in the policy-gradient path; weights $w_t$ of Eq.~\ref{eq:weights} \\
Gate / routing & $\delta{=}0$; softmax temperature $\tau{=}0.5$ (effectively arg-max with binary verifiers); one candidate per criterion (the expert of the criterion's axis); probing skipped when the rollout's task reward is $\ge0.75$; probes are greedy and cached per prompt, capped at $2{,}048$ tokens; teachers never see the rubric \\
Localization & disagreement mask: bottom-$10\%$ quantile of the teacher's log-probability of the student's tokens, dilated by $\pm1$ token (coverage $2$--$17\%$) \\
Weights & $\bar\omega_j=(\min(\omega_j,5)/5)^{1/2}$ (square-root compression of the rubric weight, exponent $\gamma{=}0.5$); pitfall criteria excluded from gating \\
Schedule & $300$ steps; validation every $50$ steps ($4$ samples per prompt); checkpoints every $25$ steps; reported checkpoints selected offline (step $175$ for the 4B run) \\
Evaluation & greedy decoding, $4{,}096$ new tokens, non-thinking, one local harness, serial judging (App.~\ref{app:benchmarks}) \\
Hardware & $8\times40$\,GB GPUs: judge server ($2$), expert sidecar ($1$), training ($3$); ${\approx}147$\,s per step for the 4B run including probes and teacher forwards (App.~\ref{app:cost}) \\
\bottomrule
\end{tabular}
\end{table}

\paragraph{Starting point.}
We ran the full pipeline from three starts of the same 4B family: a cold-start base with a short SFT stage, the
instruction-tuned hybrid model decoded without thinking (the main-text setting), and its thinking-mode sibling.
The thinking-mode start \emph{saturates}: GRPO training of the expert on the strongest axis stays flat at its zero point
($0.585$--$0.588$) for $100$ steps under two learning rates, so no expert can be trained that outscores the student under the criterion
verifiers, and the gate correctly licenses almost nothing. The non-thinking start trains ($0.541\to0.666$ on
the same axis in $200$ steps) and is used throughout the main text.

\paragraph{Checkpoint selection.}
In-training validation (sampled, mean of four) reordered the final ranking of candidate checkpoints in most
arms; every reported checkpoint is the offline winner among the top-$k$ in-training candidates under the
canonical greedy protocol, selected on the internal dev set before external evaluation.

\subsection{Compute Cost}
\label{app:cost}
\paragraph{Serving layout (8$\times$40\,GB).}
One judge server (large MoE, W8A16) on two GPUs; one \emph{sidecar} serving
the shared base plus all expert LoRA adapters on one GPU --- probes, repairs,
and teacher log-probabilities for every expert come from this single
deployment (multi-LoRA); training occupies the remaining GPUs. Teacher scoring returns the per-position log-probability of the student's
sampled token from one forward pass (a top-$K$ distribution mode exists but is
not used in the reported runs), providing both the distillation target and
the disagreement mask.

\paragraph{Cost accounting.}
Per prompt, VG-OPD adds at most $|K(j)|$ cached greedy probe generations, one forward per licensed
teacher over the response, and the verifier calls for failed criteria. Probes are capped
in length and reused across the group's rollouts.
At our scale the full pipeline runs at a median 146.8\,s/step versus 94.7\,s for
the RL-only floor ($+55\%$); the difference is concentrated in the generation
phase where probes and teacher forwards overlap rollouts (100.0 vs 56.3\,s),
while the update phase is nearly unchanged (26.0 vs 19.8\,s). MOPD and CriPO
arms run at 70--104\,s/step (no probe stage).

\subsection{Baselines}
\label{app:baselines}

\paragraph{Fairness protocol.}
All distillation-based baselines share the student, data mixture, rollout and
step budgets, and optimizer settings with VG-OPD, and differ only in how the
teacher signal is constructed (who teaches, where, with what weights). Each
baseline row uses the recipe's \emph{standard published form}; MOPD and
OPSD are pure distillation stages, as in their sources. All table rows
are decoded with the same budget, scored by one local harness with serial
judging, and selected offline (Apps.~\ref{app:benchmarks} and~\ref{app:training}).

\paragraph{GRPO floor.}
The RL-only floor shares data and budget but disables distillation. One
honest asymmetry: the floor uses the \emph{stable-optimal} RL configuration
(no entropy bonus) rather than a copy of our hyperparameters --- with the
entropy bonus that the distillation arms tolerate, pure GRPO's entropy diverges
because it lacks the mode-seeking KL term that anchors
it. Copying our config into the floor would sandbag it. Note the floor is itself
a published-method row: its reward is the graded rubric score, i.e., GRPO with
Rubrics-as-Rewards \citep{gunjal2025rubricsrewardsreinforcementlearning}, and the table marks it as such.

\paragraph{MOPD \citep{ma2026mopdmultiteacheronpolicydistillation}.}
No official code is released; we implement the paper's policy-gradient form,
the sampled-token reverse-KL estimator (their Eq.~4; the paper also reports a
top-$k$ distillation variant), with the
per-token advantage $-\hat k_t$ of Eq.~\ref{eq:opd} that every distillation
arm in our harness shares (their clipped log-ratio is the $k_1$ estimator of
the same reverse KL), prompt-level deterministic routing by the task's label
(our capability label plays the domain role), uniform token weights, and no
verification. Teachers are our four experts --- the same pool VG-OPD uses ---
so the comparison isolates the merging recipe, not teacher quality. We report
the pure form (their Stage-3: distillation only).

\paragraph{OPSD (privileged self-distillation) \citep{gu2026rethinkingrewardsupervisionrubricconditioned,bablani2026rubricsprivilegedinformationopenended}.}
The teacher is the \emph{student's own base} conditioned on privileged
context: the instance rubric appended to the user message. Teacher
log-probabilities are re-indexed onto the student's unprivileged context
position-by-position; weights are uniform. This isolates
``rubric as privileged information'' from ``verifier-licensed experts''.

\paragraph{CriPO \citep{xia2026cripoenhancingrubricbasedrl}.}
We reproduce CriPO: group-level suppressed criteria (their Eq.~4,
positive-weight criteria only); a counterfactual self-teacher --- the
\emph{live} policy served with the criterion removed from context --- scored
via per-token \texttt{prompt\_logprobs}; token localization by their Eq.~8
($\Delta>0\wedge p_T<\alpha\max_v p_T$, $\alpha{=}0.1$); and advantage
flipping by their Eq.~9 with $\tau_{\mathrm{flip}}{=}0.1$, applied only where
the true advantage is negative. The counterfactual context doubles sequence
length, so this arm trains with an enlarged context window.

\paragraph{Teacher-anchored GRPO \citep{ramos2026recipelongcontextreasoninglarge}.}
\citet{ramos2026recipelongcontextreasoninglarge} combine GRPO with on-policy
distillation by replacing the reference-policy KL of GRPO with a per-token
reverse KL to a stronger teacher, weighted by $\beta$ (their Eq.~6): outcome
rewards drive the update while the teacher supplies dense token-level
regularization. Our setting has no stronger external teacher by construction,
so the row instantiates the recipe with the strongest available privileged
teacher --- the shared base conditioned on the instance rubric (the OPSD
teacher), whose dense term is added to the shared GRPO objective of
App.~\ref{app:training} at matched $\lambda$ --- and we state this
substitution. It isolates ``dense privileged fusion'' from ours' gated,
localized fusion.

\subsection{Benchmark Details}
\label{app:benchmarks}
\paragraph{Evaluation harness.}
Every row of every table is produced by one local harness: greedy decoding, non-thinking mode, at most $4{,}096$
new tokens (an $8$k budget gave no gain for any arm), identical prompts per benchmark, and one scorer per answer
type --- letter match for multiple choice, symbolic equivalence for expressions, unit-aware numeric matching with
relative tolerance for numeric answers, and the calibrated judge for rubric items. Thinking segments, if any, are
stripped before scoring; an unclosed thinking segment scores zero. Rubric benchmarks are judged \emph{serially}:
concurrent judging under a per-call time budget deflates rubric scores by up to $0.1$--$0.2$ and we re-scored
every checkpoint serially. We re-run every checkpoint locally because the same checkpoint scored several
points apart across public harnesses, and merged LoRA artifacts are verified by a weights-differ-from-base
assertion before evaluation.

\paragraph{Benchmarks.}
Seven public benchmarks in three groups. \emph{Scientific reasoning}: GPQA-Diamond ($n{=}198$),
graduate-level multiple-choice questions in biology, physics, and chemistry written by domain experts to
resist answering by search; RaR-Science ($n{=}500$), a held-out test subset of open-ended science questions
from the Rubrics-as-Rewards collection, each scored against its own rubric by the calibrated judge; SciBench
($n{=}580$), college-level physics, chemistry, and mathematics problems from textbooks with free-form numeric
answers. \emph{Domain science}: ChemBench ($n{=}1{,}391$), chemistry questions across subfields such as general,
organic, inorganic, physical, and analytical chemistry, $1{,}148$ multiple-choice and $243$ numeric, rule-scored;
RaR-Med ($n{=}150$), open-ended medical questions from the Rubrics-as-Rewards medicine collection, rubric-judged.
\emph{General reasoning}: MMLU-Pro (a fixed subset, $n{=}2{,}000$), ten-option multiple-choice questions across
fourteen disciplines with a reasoning-heavy design; MATH500 ($n{=}500$), competition mathematics problems with
exact symbolic or numeric answers, scored by symbolic equivalence.

\paragraph{Decontamination.}
RaR-Science shares an upstream source with our training pool: we removed $74$ test items that overlapped
training derivatives and identified a $1{,}123$-item overlap channel between the RaR training split and our
pool before any evaluation; all other benchmarks were checked by exact and prefix matching.
\section{Additional Results}
\label{app:results}

\subsection{Training-Paradigm and Ablation Comparisons}
\label{app:comparisons}
\begin{figure}[h]
\centering
\includegraphics[width=\linewidth]{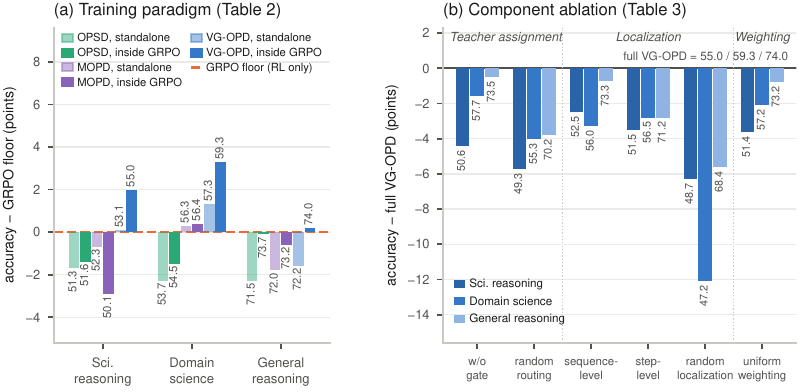}
\caption{\textbf{Tables~\ref{tab:paradigm} and~\ref{tab:component} as differences.} (a) Group-average accuracy
of each distillation recipe relative to the RL-only GRPO floor (dashed line), standalone (light) and inside GRPO
(solid); every bar is labelled with its absolute accuracy from Table~\ref{tab:paradigm}. (b) Each ablation variant
relative to full VG-OPD, one bar per benchmark group, labelled with the absolute accuracy from
Table~\ref{tab:component}.}
\label{fig:apptab23}
\end{figure}
Fig.~\ref{fig:apptab23} plots the group averages of Tables~\ref{tab:paradigm} and~\ref{tab:component} as
differences so that the two comparisons of \S\ref{sec:ablation} can be read at a glance. In panel~(a), every
standalone recipe sits below the GRPO floor on general reasoning and only VG-OPD reaches it on the two science
groups; inside GRPO, the dense recipes (OPSD, MOPD) stay at or below the floor on every group, whereas
GRPO$+$VG-OPD is the only bar above it on all three. In panel~(b), the coverage-matched random mask is the
largest drop on every group and the only variant that loses more than $10$ points anywhere ($-12.1$ on domain
science); the teacher-assignment variants (gate, routing) cost most on scientific reasoning; and flat weights
cost least, but still $3.6$ points on scientific reasoning.

\subsection{Case Studies}
\label{app:cases}

\begin{figure}[t]
\centering
\includegraphics[width=\linewidth]{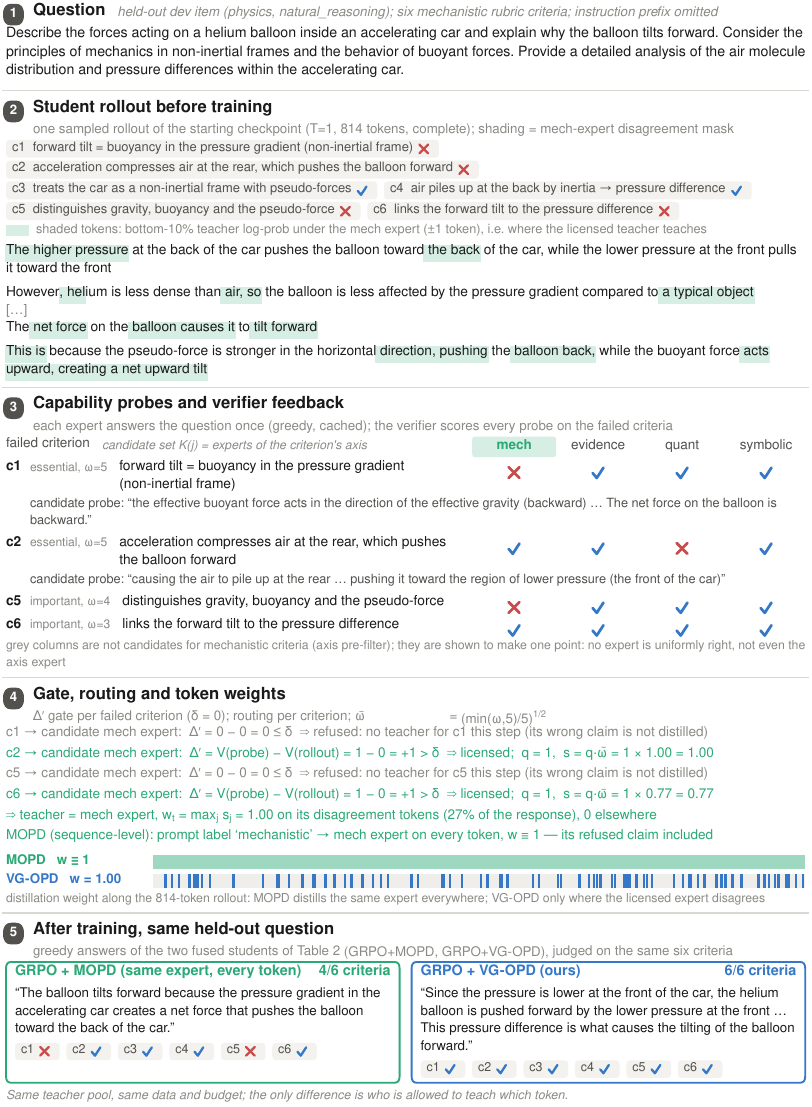}
\caption{\textbf{Case study on a held-out physics question (internal dev set).}
The domain expert is locally wrong: its probe passes the pressure-gradient criteria (c2, c6) but
claims a backward net force (c1) and fails c5. VG-OPD licenses it only where the verifier shows
it beats the rollout (c2, c6), refuses it on c1 and c5, and distills the disagreement tokens at
the verified weight; the resulting student passes all six criteria. MOPD distills the same
expert on every token; its student fails exactly c1 and c5 on this question. Shading in panel~2 is
the training-time disagreement mask under the licensed expert; every quotation is verbatim; criterion
labels are abridged (full rubric below); ``?'' marks a verifier abstention, which never gates.}
\label{fig:case2}
\end{figure}

\begin{figure}[t]
\centering
\includegraphics[width=\linewidth]{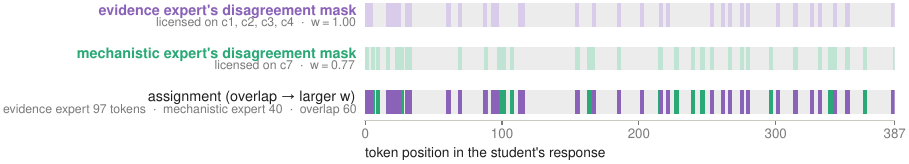}
\caption{\textbf{Token-level assignment with two licensed teachers.} A held-out medicine rollout ($387$ tokens) fails all five positive criteria; the evidence expert is licensed on c1--c4 ($w{=}1$) and the mechanistic expert on c7 ($w{=}0.77$). Each expert's disagreement mask is computed under its own log-probabilities (top two strips); tokens in both masks go to the larger verified weight (bottom), leaving disjoint token sets, each scored by its own teacher.}
\label{fig:split}
\end{figure}

One held-out physics question traces the whole mechanism end to end (Fig.~\ref{fig:case2}),
and a held-out medicine rollout shows the token-level assignment when two experts are licensed
(Fig.~\ref{fig:split}). All texts are cached greedy
outputs of the models in Tables~\ref{tab:main} and~\ref{tab:paradigm}, all verdicts come from
the verifier used in training, and the rollout of Fig.~\ref{fig:case2} is one of four samples
drawn at $T{=}1$ from the starting checkpoint (two of the four already pass every criterion, so
the case study shows a failing sample that training acts on). Questions were selected for the
clearest teacher disagreement.

\paragraph{Physics (Fig.~\ref{fig:case2}): a helium balloon in an accelerating car.}
\textbf{(1)} The rollout states that the higher rear pressure pushes the balloon \emph{toward
the back}, that helium is ``less affected by the pressure gradient'', and that the forward tilt
comes from the pseudo-force plus buoyancy; it fails c1, c2, c5 and c6.
\textbf{(2)} Probes: the mechanistic expert --- the only candidate for these mechanistic
criteria --- explains the pressure gradient correctly (c2, c6 pass) but gets the tilt itself
wrong: ``the effective buoyant force acts in the direction of the effective gravity (backward)
\ldots\ The net force on the balloon is backward'' (c1 fail), and it fails c5. The other three
experts pass c1 but are not candidates.
\textbf{(3)} Gate: $\Delta'=0$ on c1 and c5 (refused; no teacher for those criteria this
step), $\Delta'=+1$ on c2 and c6 (licensed, $s=1$ and $0.77$); the teacher is the
mechanistic expert with $w_t=1$ on its disagreement tokens ($27\%$ of the response). MOPD
distills the same expert on all $814$ tokens at $w\equiv1$, refused claim included.
\textbf{(4)} After training: the GRPO$+$MOPD student fails exactly c1 and c5, writing that the
pressure gradient ``creates a net force that pushes the balloon toward the back of the car'';
the GRPO$+$VG-OPD student passes all six, explaining that the balloon ``is pushed forward by the
lower pressure at the front''.

\paragraph{Token-level assignment with two licensed teachers (Fig.~\ref{fig:split}).}
When a rollout's failed criteria license two experts, each expert's
disagreement mask is computed under its own log-probabilities
(Eq.~\ref{eq:mask}) and a token in both masks goes to the larger verified
weight (Eq.~\ref{eq:weights}). On the medicine rollout of
Fig.~\ref{fig:split} the two masks overlap on $60$ tokens, because the experts
are adapters on one base and find largely the same student tokens unlikely;
the assignment leaves the evidence expert $97$ tokens and the mechanistic
expert $40$. The second teacher's exclusive share is therefore modest, and the
per-criterion split acts mostly through which criteria license which expert.

\end{document}